\documentclass[10pt, conference]{IEEEtran}
\usepackage{graphicx}
\usepackage{amsmath}
\usepackage{amssymb}
\usepackage{booktabs}
\usepackage{algorithm}
\usepackage{algpseudocode}
\usepackage{subfigure}
\usepackage[backend=biber, style=ieee]{biblatex}
\usepackage[breaklinks]{hyperref}
\usepackage[capitalize]{cleveref}
\crefname{section}{Sec.}{Secs.}
\Crefname{section}{Section}{Sections}
\Crefname{table}{Table}{Tables}
\crefname{table}{Tab.}{Tabs.}

\IEEEoverridecommandlockouts
\newcommand{\model}{\textbf{\texttt{NasHD}}}

\begin{document}

\title{NasHD: Efficient ViT Architecture Performance Ranking using Hyperdimensional Computing}

\author{Dongning Ma\\
Villanova University\\
Villanova, PA 19085\\
{\tt\small dma2@villanova.edu}
\and
Pengfei Zhao\\
Beijing Xiaochuan Technology Co., Ltd.\\
Haidian, Beijing 100191, China\\
{\tt\small zhaopengfei2014@xiaochuankeji.cn}
\and
Xun Jiao\\
Villanova University\\
Villanova, PA 19085\\
{\tt\small xun.jiao@villanova.edu}
}
\maketitle


\begin{abstract}
Neural Architecture Search (NAS) is an automated architecture engineering method for deep learning design automation, which serves as an alternative to the manual and error-prone process of model development,  selection, evaluation and performance estimation. However, one major obstacle of NAS is the extremely demanding computation resource requirements and time-consuming iterations particularly when the dataset scales. In this paper, targeting at the emerging vision transformer (ViT), we present \model, a hyperdimensional computing based supervised learning model to rank the performance given the architectures and configurations. Different from other learning based methods, \model ~is faster thanks to the high parallel processing of HDC architecture. We also evaluated two HDC encoding schemes: Gram-based and Record-based of \model ~on their performance and efficiency. On the VIMER-UFO benchmark dataset of 8 applications from a diverse range of domains, \model ~Record can rank the performance of nearly 100K vision transformer models with about 1 minute while still achieving comparable results with sophisticated models. 
\end{abstract}

\section{Introduction}
\label{sec:intro}

\begin{figure}
  \centering
  \begin{subfigure}{}
    \includegraphics[width = .7\columnwidth]{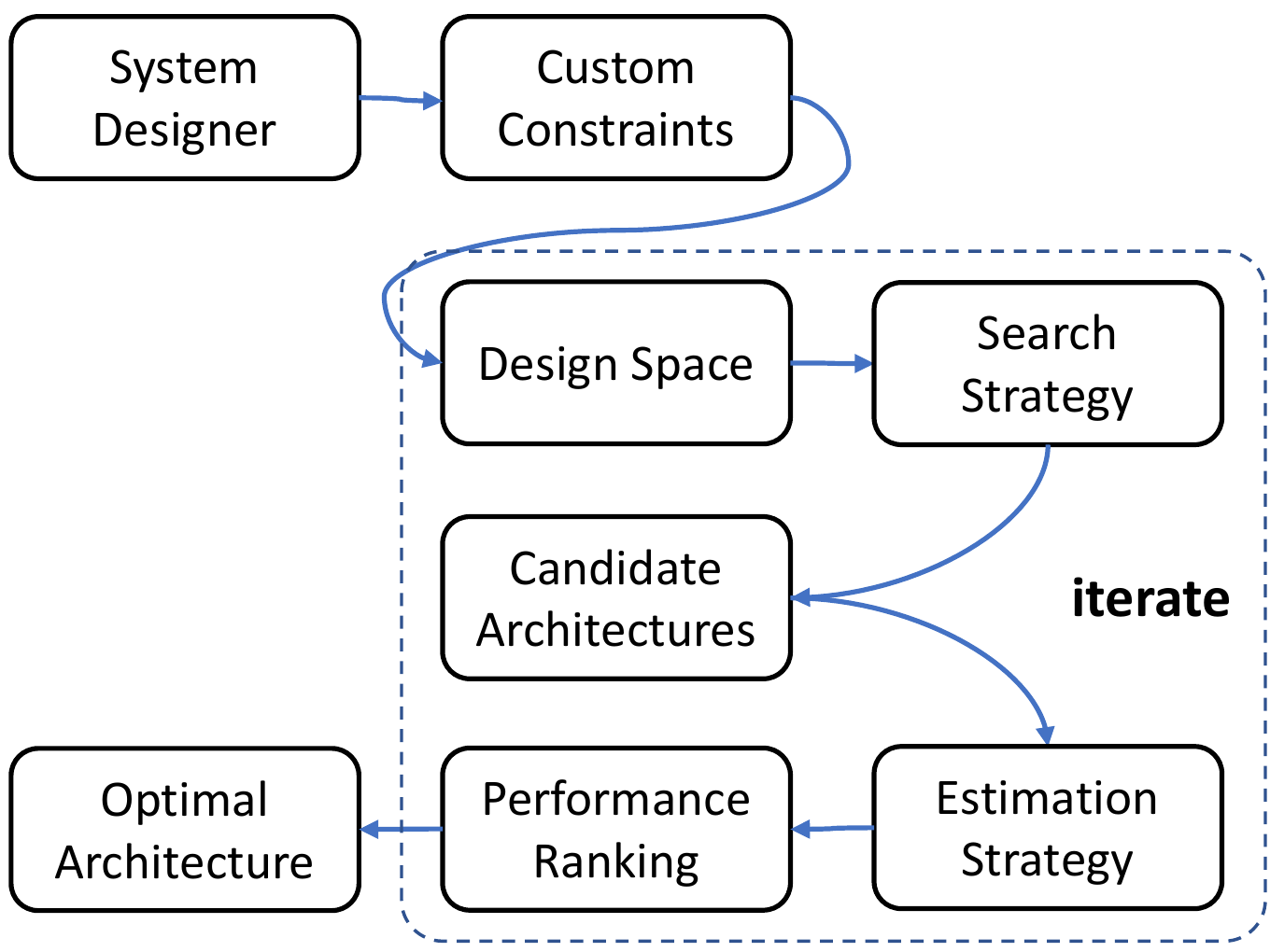}
  \end{subfigure}
  \hfill
  \caption{A typical NAS flow.}
  \label{fig:nas}
\end{figure}

As machine learning broadens its application and enhances its capability to various domains, there emerges a ever growing demand of developing and optimizing architectures for higher model performance. However, with the architectures evolving increasingly deep, the architecture engineering usually requires enormous effort due to the expansion of design space, which is hardly any possible using manual effort. To mitigate such challenge, neural architecture search (NAS), subsequently emerges as an automated process of developing and evaluating architecture given a set of customized constraints~\cite{liu2021survey, ren2021comprehensive}. In general, NAS can be summarized as \cref{fig:nas}: First, system designers set custom constraints which defines the search space. Then within this search space, candidate architectures are sampled using search strategies iteratively. Their performance is also evaluated (estimated) to obtain a ranking which determines the optimal architecture for this NAS task.

Deep neural networks (DNN), are the primary targets of neural architecture search (NAS) given its strong capability on learning tasks as well as the deep search space of its architectures. Recently, vision transformer (ViT) emerges as another promising machine learning algorithm which achieves even higher performance on computer vision tasks than some state of the art neural networks~\cite{dosovitskiy2020image}. NAS becomes even more useful and also demanding for ViT because of the heterogeneity of components and blocks as well as the corresponding configurations inside which enables even more knobs or optimization. On the other hand, the huge amount of computation resources spent on NAS process also becomes a major concern, e.g., clusters of hundreds of GPUs usually spend days to weeks to obtain architectures that can achieve comparable results on more challenging computer vision tasks~\cite{tan2019mnasnet}.

One major computation-expensive process in NAS is to estimate the performance of architectures sampled from the predefined search space. A canonical training and validation flow can be applied to estimate the performance of an architecture. However, this is extremely slow thus much recent methodologies rely on various speed-up techniques such as sharing weights from a supernet to avoid training models afresh~\cite{xie2021weight, chen2021neural, yang2021towards}. Although achieving acceleration, it is still necessary to compare and rank the performance of the sampled architectures for selection, which motivates the use of machine learning to rank the performance of sampled architectures without actual training or inference. 

In this work, we depart from the conventional learning algorithms and embrace a novel and brain inspired algorithm -- hyperdimensional computing (HDC) to achieve even more efficient performance ranking of the ViT architecture performance. HDC is an emerging non von Neumann computation scheme, leveraging the learning and representation capabilities of extremely high dimensional vectors inspired from the abstract brain activity functionalities~\cite{thomas2021theoretical}. As a learning algorithm, HDC has achieved advantage over conventional machine learning algorithms such as neural networks for its high energy efficiency, smaller model size and strong support for parallel computing~\cite{hassan2021hyper}. In this paper, we introduce this emerging computing scheme into the domain of NAS by presenting \model, a supervised vision transformer performance ranking algorithm using HDC. \model ~can efficiently rank the performance of architectures given the configurations of vision transformers. To the best of our knowledge, this is the first work to leverage HDC for NAS. The main contributions of this paper are as follows:
\begin{itemize}
    \item We introduce the emerging brain inspired HDC into the domain of NAS. Specifically, we develop a model that can rank the performance given a set of vision transformer architectures without actual training. 
    \item We propose two different HDC encoding schemes (Gram and Record) according to the architecture of vision transformers and evaluate and compare their performance. We also propose retraining methods based on weight-update to enhance model performance after initial training.
    \item We evaluate \model ~performance on the VIMER-UFO benchmark of 8 computer vision applications from different domains and compare with various baselines. Experimental results show that \model ~can rank around 100K vision transformer architectures with around 1 minute while still achieving comparable results.
\end{itemize}

\section{Related Works}
\label{sec:relate}
\subsection{Neural Architecture Search of Vision Transformers}
As ViT is a relatively new architecture compared with DNNs, the NAS strategies for ViT are usually referred from NAS for DNNs. For example, Supernet is generally used for SOTA NAS frameworks and also applied in NAS for ViTs. NASViT~\cite{gong2021nasvit} targets at the obstacles on introducing supernet training from DNNs to ViTs, and propose techniques including gradient projection algorithms, switchable layer scaling designs, simplified data augmentations and regularization training recipes. Autoformer~\cite{chen2021autoformer} also proposes weight entanglement specific to the ViT architectures which possess advantages such as fast convergence, lower memory overhead and better sub-net performance compared to conventional weight sharing methods. NAS of ViT can also have more specific architecture orientations, e.g., VTCAS~\cite{zhang2022vision} considers both the ViT as well as the desirable properties of convolutional architectures during NAS, which achieves 82.0\% top-1 accuracy on ImageNet dataset.

\subsection{Hyperdimensional Computing}
HDC leverages the learning capability of high dimensional numerical vectors to perform tasks such as classification and regression~\cite{ge2020classification, hernandez2021reghd}. HDC is also used to detect anomalies or outliers, for example, to identify potential attacks to vulnerable sensors~\cite{wang2021brief}. For natural language processing, HDC can also achieve comparable accuracy of other machine learning models such as kNN and SVM with much smaller model size and less computation overhead~\cite{thapa2021spamhd, liu2022l3e}. Another outstanding advantage of HDC is its strong compatibility on heterogeneous platforms, for example, various acceleration techniques for HDC are implemented in GPGPUs~\cite{kang2022xcelhd}, FPGAs~\cite{salamat2019f5} and even the emerging in-memory computing architectures~\cite{liu2019hdc}. HDC also shows stronger robustness against hardware errors from voltage scaling or memory failures, which enables it to be deployed on such approximate computing platforms~\cite{zhang2021assessing, zhang2022energy}. NAS for HDC has been recently discussed to find optimal HDC architectures for a specific task~\cite{yang2022automated}, however using HDC for NAS is still at large.

\section{Hyperdimensional Computing}
\label{sec:prelim}

\subsection{Notions and Operations}
\label{sec:notions}
\textbf{Hypervector} Hypervectors (HV) are the fundamental elements of an HDC model and are also the minimal units of operations. HVs are numerical vectors with several specific characteristics. 1). \textbf{high-dimensional}: the dimension (the amount of numbers) of HVs are extremely high, usually reaching 10,000 and above; 2). \textbf{holographic}: each HV is recognized as the minimal unit in an HDC model, individual number inside HV does not own unique individual representation; 3). \textbf{(pseudo-)random}: for the initial generated HVs, the numbers are i.i.d., thus two randomly generated HVs are approximately orthogonal to each other due to the extremely high dimensionality. We use $\vec{V} = (v_1, v_2, ... , v_D)$ as the notion of a $D$-dimensional HV in which $v_i$ is the $i$-th number inside the HV. For clarification purposes, HVs are marked with arrows in this paper, while other vectors such as feature vectors are bold instead.

\textbf{HV Operations} While an HV can be used to represent information, in reality the information can come from diverse modalities, and different sources of information can correlate. To aggregate information of the same modality or to combine information from different modalities to also provide hierarchy of information, we use HDC operations. The three mostly used HDC operations are addition, multiplication and permutation. The addition and multiplication operations take two HVs as inputs and perform linear and element-wise vector operations correspondingly. The permutation operation takes only one HV as input and perform cyclic shift of a specific amount $n$. Note that the dimensions of input HVs and output HVs are the same for all the three operations.

Bipolarization (or binarization) is another important operation for HV beyond the three basic operations. Bipolarization takes the sign bit of the HV elements that any number larger than $0$ is bipolarized into $1$, while any number smaller than $0$ is bipolarized into $-1$. Binarization is similar, but numbers are binarized into $1$ and $0$ instead. Bipolarization and binarization introduces additional non-linearity into HDC models, and also to limit the range of elements inside each HV to prevent overflow issues during aggregation. In \model, we slightly modify the bipolarization behavior that any number larger than $1$ or smaller than $-1$ will be capped into $1$ and $-1$ while the numbers within $(-1, 1)$ are kept as-is.

\textbf{Similarity}
With HVs representing information and HDC operations aggregating and combining information, there is then a need of metric that can quantitatively measure the similarity between information that different HVs accommodate. Cosine similarity is one of the most frequently used metric. A higher cosine similarity indicates that the two HVs compared share more similar information, thus are more alike. 

\subsection{HDC Memories}

HDC leverages HDC memories to host information. HDC memories are a specialized cluster of HVs with different objectives during a learning task. There are two major categories of HDC memories: item memory and associative memory. Item memories are related with the input realistic features: each item memory hosts item HVs at the same amount of possible feature values $K$ of the corresponding features as shown by $\mathbb{I} = \{\vec{I_1}, \vec{I_2}, ... \vec{I_K} \}$. If the feature is a continuous variable, quantization can be applied beforehand to avoid item memories of infinite size. Associative memory, on the other hand, is related to the output of the model. For a classification task, the associative memory hosts class HVs, each of which represents a class of the task, as shown by $\mathbb{A} = \{\vec{A_1}, \vec{A_2}, ... \vec{A_T} \}$. For this paper, since we are ranking the performance of various ViT architectures, the associative memory of \model ~hosts an HV aiming to represent high performance architectures. Details of how the associative memory can help with ranking architecture performance are present in \cref{sec:model}.

\section{\model ~Methodology}
\label{sec:model}

\subsection{Problem Formulation and Motivation}
In NAS, architectures are usually sampled from a supernet. Designers usually have a small set of architectures with their performance ranked, while have a much larger set to rank. Could we develop an algorithm or model that can efficiently train on this small set of labeled architectures and accurately rank a significantly larger set without training them individually? We found that two highlighted characteristics of HDC can potentially enable a new direction of addressing this problem: 1. HDC is known for its efficiency over existing machine learning algorithms such as neural networks, including smaller model size, faster model convergence and less computational intensity~\cite{kim2020geniehd, thapa2021spamhd, ma2021molehd}. 2. HDC is better at learning with limited data such as one-shot or few-shot learning tasks and are less likely to over-fit~\cite{burrello2018one, rahimi2018efficient}. 

\subsection{Overview of \model}
\begin{figure*}
  \centering
  \begin{subfigure}{}
    \includegraphics[width = 1.64\columnwidth]{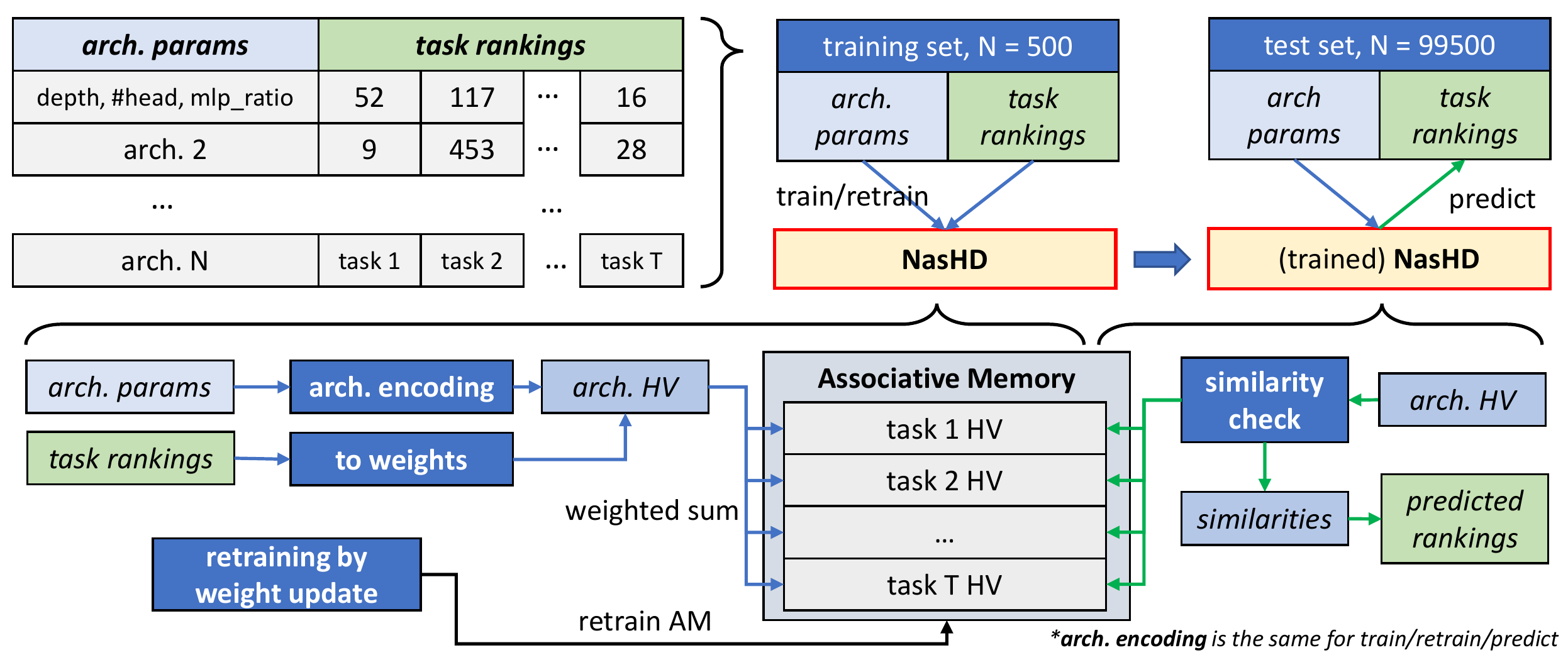}
  \end{subfigure}
  \hfill
  \caption{Train \model ~to rank ViT performance based on architecture.}
  \label{fig:nasvithd}
\end{figure*}

An overview of using \model ~to learn the features of a small training set and to rank the architecture performance a large testing set is present in \cref{fig:nasvithd}. \model ~first iterates through the training set, takes parameters of all the architectures and encodes them into the corresponding HVs referred to as the architecture HVs. \model ~also converts the task rankings into weights of each architecture in the training set. Then, \model ~uses the weights to perform weighted sum to establish the associative memory which accommodates the task HVs. In brief, architectures with higher rankings are assigned with larger weights thus having more information impact on the task HVs after the weighted sum. 

When predicting rankings of architectures from the test set, \model ~performs the same architecture encoding to obtain the architecture HV. Then \model ~computes and checks the similarity between the HV and each task HV in the associative memory. Once the similarity metrics of all the architectures from the test set are obtained, \model ~can obtain the ranking of the similarity metrics as the predicted rankings. To enhance \model ~performance, we also propose ``retraining by weight-difference'' and update the associative memory accordingly.

\subsection{Dataset Description}
\label{sec:dataset}
The available training data consists of three architectural parameters including the depth of encoders ($depth$), number of attention heads ($\#head$) and the dilation ratio of the MLP ($mlp\_ratio$) of each layer. Each parameter has three possible values: $depth$: $\{10, 11, 12\}$; $\#head$: $\{10, 11, 12\}$; $mlp\_ratio$: $\{3.0, 3.5, 4.0\}$. Therefore, for each encoder, there are $3\times3$ possible parameter combinations, so the number of possible architectures in the search space are $(3\times3)^{10} + (3\times3)^{11} + (3\times3)^{12}$. For convenience in data processing, the parameters are (label-)encoded into numbers of $\{1, 2, 3\}$. 

The labels are the the performance rankings of the architecture on different tasks. Smaller number refers to higher performance, e.g., ranking of $0$ means this architecture is the top performance architecture of this task. 

\subsection{ViT Architecture Encoding}
\begin{figure*}
  \centering
  \begin{subfigure}{}
    \includegraphics[width = 1.64\columnwidth]{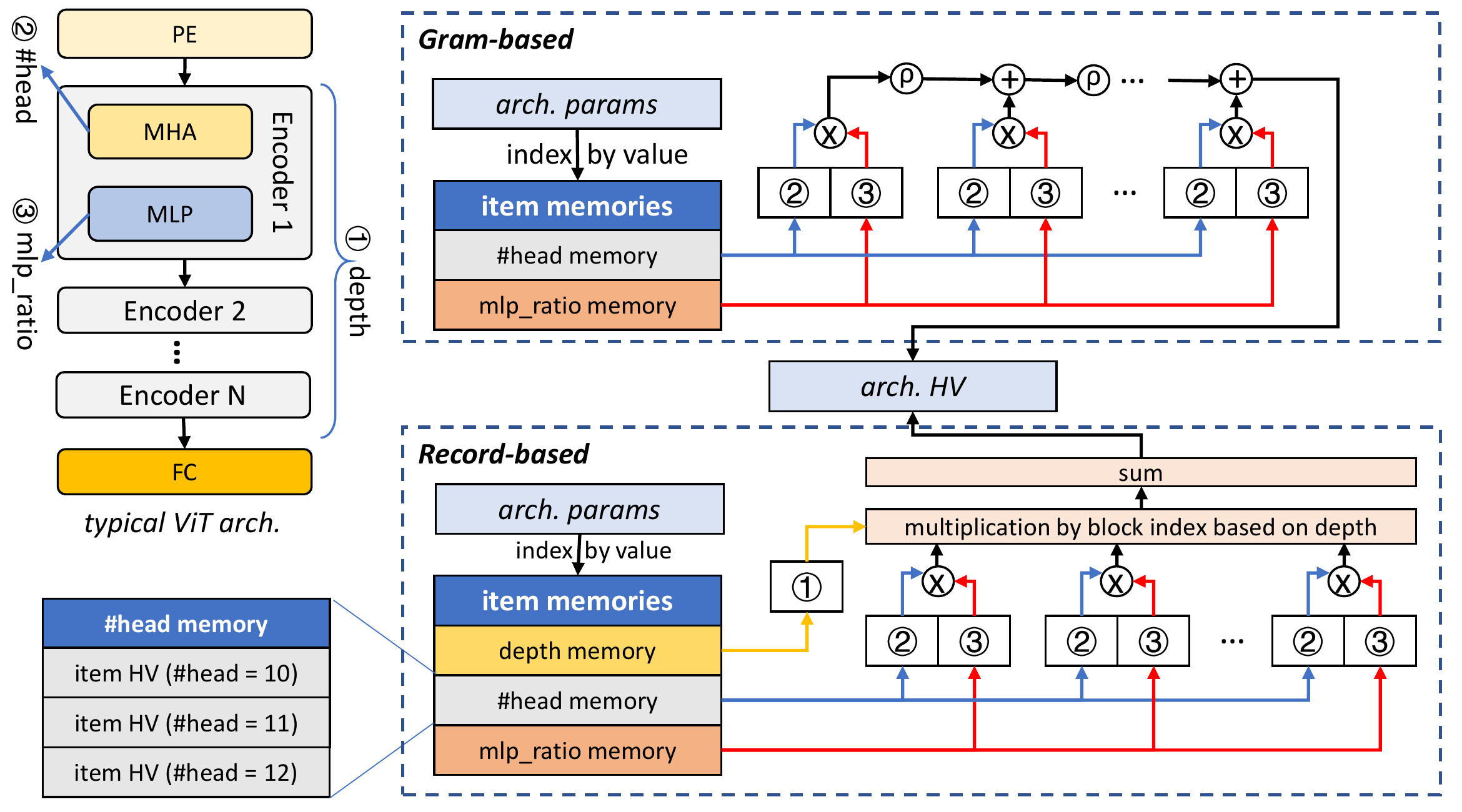}
  \end{subfigure}
  \hfill
  \caption{Encoding a ViT architecture into an HV. Two schemes are introduced: Gram-based and Record-based encoding.}
  \label{fig:encode}
\end{figure*}

In this subsection, we introduce the architecture encoding in detail. Specifically, we propose two schemes of encoding: Gram-based encoding and Record-based encoding, as indicated by \cref{fig:encode}. Note that the architecture HVs $\vec{V}$ are initialized with $0$.

\subsubsection{Gram-based Encoding}
Gram-based encoding recognizes each encoder block as a ``gram'' and groups parameters by tuples based on which encoder they belong to. It features two item memories: the $\#head$ memory and the $mlp\_ratio$ memory. Since each parameter has three possible values as introduced in \cref{sec:dataset}, each item memory hosts three item HVs, each representing a possible value in the high-dimensional space.

Gram-based encoding first obtains the corresponding item HVs from the item memory based on the parameter tuples of $\#head$ and $mlp\_ratio$. The two indexed HVs are aggregated by HV multiplication, resulting the gram HV. Then, the gram HVs are permuted where the shift amount is based on the depth index of the encoder, i.e., encoders closer to the output of the ViT model are shifted for more dimensions, or vice versa. The permuted gram HVs are summed up into the architecture HV $\vec{V}$. Encoded HVs are bipolarized (or binarized). 

Note that the Gram-based encoding can be paralleled, since the encoding within each gram is independent from each other, and the shift amount of permutation is solely dependent on the depth index. A major disadvantage of Gram-based encoding, however, is the permutation operation is not very straightforward for implementation of acceleration, since the cyclic rotation of high dimensional vectors are not linear vector operations like addition and multiplication. Therefore, Record-based encoding is an alternative to the Gram-based encoding which eliminates the use of permutation, but only keeps addition and multiplication of HVs.

\subsubsection{Record-based Encoding}
Record-based encoding uses one additional item memory, the depth memory as an alternative to the gram-based permutations. The depth memory hosts item HVs at the number of depth index, therefore, there are 12 item HVs in the depth memory given the max possible depth is 12 for this dataset. Within each encoder, the HV aggregation is the same as Gram-based encoding. However, instead of permutation by the depth index, the HV of each encoder is multiplied by the item HV of the corresponding depth, indexed from the depth memory. Then the HVs are summed up to obtain the architecture HV, which is also bipolarized (or binarized). In realistic implementation, based on the commutative properties of multiplication, the item HVs in the item memories can encode first and stored as a unified item memory to avoid repetitive index and HV multiplications. 

\subsection{\model ~Training}
Training is the process to establish the associative memory which accommodates the task HVs, which are initialized with all elements to be 0. In \model, training is the weighted sum of the HVs obtained from the encoding phase as described in \cref{eq:training}, where $\vec{A_t}$ is the HV of task $t$ in the associative memory, $w^{(t)}_n$ is the task $t$ weight of the $n$-th architecture in the training set, and $ \vec{V_n}$ is the encoded HV of the $n$-th architecture as well. Architectures with higher rank (smaller in number) of a specific task are assigned with higher weights, while architectures with lower rankings (larger in number) are assigned with lower weights. A custom learning rate $\gamma$ can be specified, however, during training we use the constant $1$. The objective of training is to incorporate more information about high performing architectures into the task HV, while still trying to incorporate information from HVs as much as possible to avoid over-fitting, since there are just 500 architectures in the training set. 

\begin{equation}
    \vec{A_t} = \vec{A_t} + \gamma \times w^{(t)}_n \times \vec{V_n} \; \; (n = 1, 2, \cdots, N)
    \label{eq:training}
\end{equation}

Intuitively, we convert the ranking into weights using a straightforward inverse number method as \cref{eq:weight}. $\boldsymbol{w_n} = \{w_{n1}, w_{n2}, \cdots, w_{nT}\}$ is the weight vector of the $n$-th architecture, in which $w_{it}$ refers to the weight of $t$-th task. $\boldsymbol{r_n} = \{r_{n1}, r_{n2}, \cdots, r_{nT}\}$ is the ranking vector from the training set, with rankings of each task correspondingly, as shown in \cref{fig:nasvithd}. $\mu$ is a constant scaling factor and we use $1.0$ during the experiments. Therefore, worse-than-average performing architectures are assigned with negative weights and better-than-average ones are assigned with positive weights.

\begin{equation}
    \boldsymbol{w_n} = \mu(\textbf{1} - \frac{\textbf{2}}{\boldsymbol{r_n} + \textbf{1}})
    \label{eq:weight}
\end{equation}

\subsection{\model ~Prediction}
For the test set, instead of directly predicting the ranking, \model ~uses the similarity metrics to determine the ranking. For an architecture from the test set with unknown ranking, \model ~first encodes its parameters into the architecture HV using the same item memories and encoding scheme for training. Then, \model ~checks the similarity between the architecture HV and the task HVs in the associative memory. For each task, the similarity of all the architectures are recorded and then used to rank the task performance. Specifically, as the task HV in the associative memory is the aggregation of architectures with high performance, therefore, a higher similarity will then be predicted with higher ranking by \model.

\subsection{\model ~Retraining}
In HDC, training usually takes one epoch (a full iteration of the entire training set). Additional epochs of retraining (still using the training set) can be optionally performed to enhance the performance of the model. HDC models for classification tasks leverages prediction labels to update the associative memory when a mis-classification is identified~\cite{thapa2021spamhd}. However, \model ~targets at a ranking task, therefore we propose a novel retraining methodology to increase the consistency of predicted rankings and the ground-truth.

\model ~retraining is based on updating the associative memory by weight-difference. The main concept is to make sure the similarity ranking of each architecture is consistent with the ground-truth ranking to minimize the Kendall tau similarity between the prediction and the ground-truth. First after the training epoch, \model ~calculates the similarity metrics of all the $N$ architectures in the training set which is referred to as $\boldsymbol{\delta} = \{\delta_1, \delta_2, \cdots, \delta_N\}$. Based on the similarity we are able to obtain the predicted rankings $\boldsymbol{\hat{r}} = \{\hat{r}_1, \hat{r}_2, \cdots, \hat{r}_N\}$, which can be subsequently converted into the speculated weights $\boldsymbol{\hat{w}} = \{\hat{w}_1, \hat{w}_2, \cdots, \hat{w}_N\}$ according to \cref{eq:weight}.

We use the difference between the speculated weights and the ground-truth weights $\boldsymbol{\Delta{w}} = \{w_1 - \hat{w}_1, w_2 - \hat{w}_2, \cdots, w_N - \hat{w}_N\}$ to perform training again to update the associative memory based on \cref{eq:training}. A negative weight difference means the architecture is ``under-ranked'' and its HV should be added into the task HV during retraining, on the other hand, a positive difference means the architecture is ``over-ranked'' and its HV should be subtracted instead. This roughly resembles the error back-propagation algorithm in conventional machine learning algorithms such as neural network. Retraining can iterate for multiple epochs and can be stopped when the (average) difference is smaller than a threshold to prevent over-fitting. We use the starting learning rate of $1$ for consistency with training, however, for additional epochs in retraining we apply a decay of $0.8$.

\section{Experimental Results}
\subsection{Experimental Setup}
The dataset of architecture performance is on the multi-tasks of VIMER-UFO benchmark~\cite{xi2022ufo}, with 8 datasets from different computer vision tasks: CPLFW~\cite{kim2020groupface}, Market1501~\cite{herzog2021lightweight}, DukeMTMC~\cite{ristani2014tracking}, MSMT-17~\cite{he2021transreid}, Veri-776~\cite{huynh2021strong}, VehicleId~\cite{he2020fastreid}, VeriWild~\cite{he2020fastreid}, and SOP~\cite{lee2020compounding}. This benchmarks has 500 architectures in the training set with performance ranked on each of the task and 99,500 for ranking prediction. We implement \model ~using PyTorch and evaluate \model ~with one Nvidia P100 GPU. We also compare \model ~with four baseline methods:
\begin{itemize}
    \item GP-NAS~\cite{li2020gp}, which stands for Gaussian process neural architecture search in which the correlation between performances and architectures and the correlation between different architectures are explicitly modeled. An efficient sampling method is also proposed which enables GP-NAS learning on a small set of samples.
    \item LightGBM~\cite{ke2017lightgbm, he2022skt}, which is a widely adopted gradient boosting decision tree.
    \item CatBoost~\cite{dorogush2018catboost}, is another gradient boosting baseline but with categorical features support.
    \item GP-NAS Ensemble~\cite{chen2022gp}, which is an ensemble version of GP-NAS~\cite{li2020gp}, it applies enhancements such as additional feature engineering, label transformation and weighted ensemble kernels.
\end{itemize}

\subsection{Ranking ViT Architectures}
We present the comparison between \model ~and the baselines on the VIMER-UFO benchmark in \cref{tab:score}. Kendall tau is used to describe the consistency between the model prediction and the ground-truth, a higher scores means the ranking predictions are more accurate. Kendall tau $\tau$ between two vectors of ranking can be calculated by \cref{eq:tau}, where $n_c$ and $n_d$ are the number of concordant and discordant pairs, respectively and $n$ is the total number of pairs~\cite{kendall1948rank}.

We can observe that the CPLFW is the most challenging task of all the 8 tasks that all the models are having significantly lower score than the rest 7 tasks, making this face related dataset the most challenging task in this benchmark. As to individual models, GP-NAS, \model ~Grams and LightGBM are the two sub-performing as their average scores are both less than 0.7. CatBoost, GP-NAS Ensemble, and \model ~Record are the top 3 models, with scores achieving over 0.785. 

\begin{equation}
    \tau = \frac{n_c - n_d}{n(n-1)/2} 
    \label{eq:tau}
\end{equation}

\begin{table*}[htbp]
  \centering
  \caption{Comparison of Models on Kendall Tau Scores of the VIMER-UFO Benchmark}
    \begin{tabular}{cccccccccc}
    \toprule
    \textbf{Model} & \textbf{Average} & \textbf{CPLFW} & \textbf{Market} & \textbf{MTMC} & \textbf{MSMT} & \textbf{Veri} & \textbf{VehicleId} & \textbf{VeriWild} & \textbf{SOP} \\
    \midrule
    GP-NAS & 0.6196 & 0.2350 & 0.7391 & 0.7052 & 0.8063 & 0.6319 & 0.4012 & 0.6731 & 0.7654 \\
    LightGBM & 0.6810 & 0.2755 & 0.7742 & 0.7723 & 0.7599 & 0.7469 & 0.5900 & 0.7860 & 0.7433 \\
    Catboost & 0.7851 & 0.3220 & 0.8687 & 0.8883 & 0.9430 & 0.8930 & 0.6576 & 0.9099 & 0.7980 \\
    GP-NAS Ensemble & 0.7978 & 0.3188 & 0.8864 & 0.9045 & 0.9678 & 0.9106 & 0.6624 & 0.9199 & 0.8119 \\
    \midrule
    \model ~Gram & 0.7608 & 0.2927 & 0.8489 & 0.8580 & 0.9004 & 0.8571 & 0.6598 & 0.8819 & 0.7874 \\
    \model ~Record & 0.7899 & 0.3074 & 0.8754 & 0.8945 & 0.9555 & 0.8974 & 0.6634 & 0.9195 & 0.8060 \\
    \bottomrule
    \end{tabular}%
  \label{tab:score}%
\end{table*}%

\subsection{Comparison on Efficiency}
We compare the run-time of \model ~and baselines to show the efficiency of \model. The execution time for training the model and making predictions are listed in \cref{tab:efficiency}. For less sophisticated models such as GP-NAS, LightGBM are fast to learn and predict, however their performance is relatively sub-par. CatBoost, based on the time reported, uses relatively short execution time. However, to achieve such enhanced performance, it requires extensive parameter tuning which also takes much longer time than just training and prediction. This parameter optimization is required for each task, which also limits the flexibility of this algorithm. GP-NAS Ensemble, the more complicated model, achieves the highest performance overall however during our evaluation, it spends way more than 10 minutes for training and prediction since each of the ensemble models require intensive training effort. We mark the execution time with $>$600 since it is already much time-consuming than most of the baselines.

The time reported for both of the \model ~implementations include the entire process of all the tasks, i.e., no additional pre-training or fine-tuning is required. \model ~Record can finish training and prediction together with around 1 minute. \model ~Gram requires much longer processing time compared to most of the baselines, as the permutation operations of HVs occupy most of the additional effort spent. As \model ~Gram also shows sub-par performance, so we conclude that \model ~Record is preferred for this benchmark.

\begin{table*}[htbp]
  \centering
  \caption{Comparison of Execution Time of the VIMER-UFO Benchmark}
    \begin{tabular}{ccccccc}
    \toprule
    \textbf{Model} & GP-NAS & LightGBM & CatBoost & GP-NAS Ensemble & \model ~Gram & \model ~Record \\
    \midrule
    \textbf{Execution Time (second)} & 92    & 69    & 80    & $>$600  & 425   & 62 \\
    \bottomrule
    \end{tabular}%
  \label{tab:efficiency}%
\end{table*}%

\subsection{Impact of HV Dimension}
\begin{figure}
  \centering
  \begin{subfigure}{}
    \includegraphics[width = .84\columnwidth]{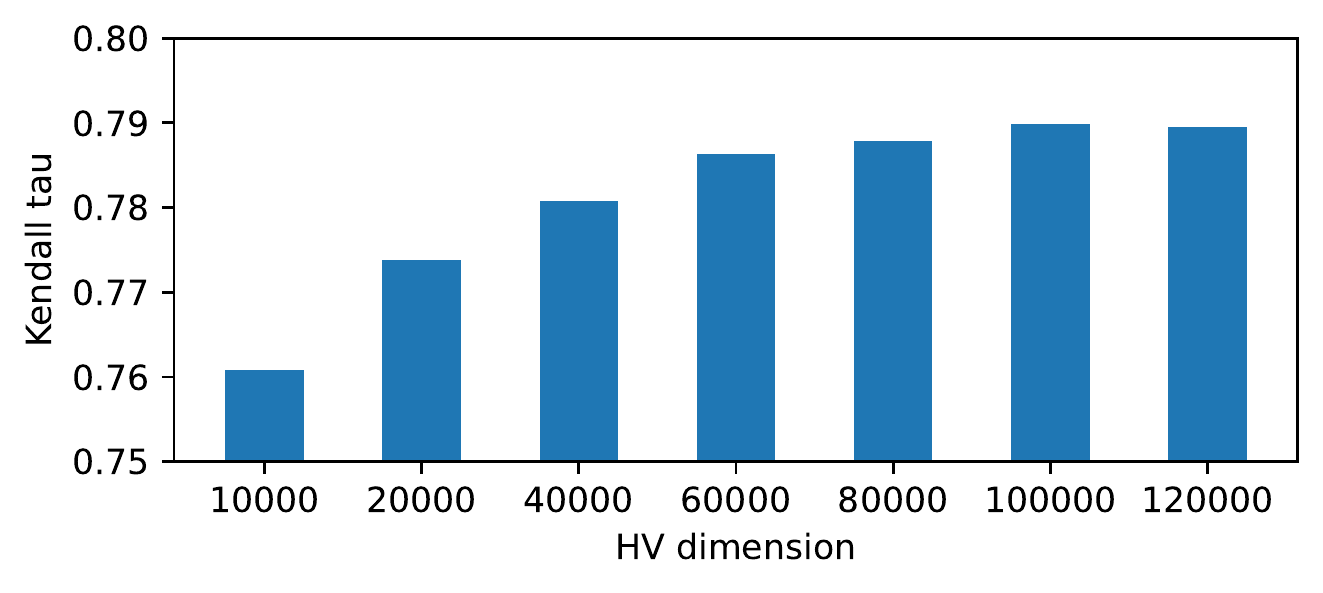}
  \end{subfigure}
  \hfill
  \caption{Task-average scores of \model ~Record under different HV dimensions.}
  \label{fig:dim}
\end{figure}

We also present a case study on the Record-based encoding to analyze the impact of using different HV dimensions on \model ~performance. In related HDC literature, dimensions of 10,000 -- 20,00 is usually used as the dimension range and higher dimensions may not guarantee a higher performance. Therefore, we choose 10,000 as the starting point of HV dimension analysis on \model. However, we observed that the score can be increased to over 100,000 without performance saturation. Experiments show that a dimension around 120,000 achieves the best performance across all the dimensions evaluated, as shown in \cref{fig:dim}. 

\subsection{Discussion}
The efficiency of \model ~comes from the simplicity of HDC training and retraining, which follows a concise linear operations with matrices (record-based encoding). This can be hugely benefited from the high parallelism of GPU processing that significantly accelerates the \model. \model ~also does not need back-propagation like neural networks, which requires maintaining a complicated computation graph. Instead, \model ~retraining is via the update to the associative memory which is also a weighted sum with much less computational overhead. Another advantage is that when evaluating different tasks, \model ~can share the same item memory and encoding process. This grants a tremendous advantage over other baselines that different tasks are required to train individual models or apply relatively more complicated transfer processing. 

\section{Conclusion}
Efficiency of neural architecture search (NAS) algorithms has been one major bottleneck for this automated model engineering process. The emerging vision transformer (ViT) algorithm puts even higher demand on NAS for its deep and complicated architectures. Specifically, how to efficiently and accurately rank a larget set of ViT model performance given a small amount of training set is a practical problem in ViT model engineering. In this paper we leverage the brain-inspired hyperdimensional computing (HDC) computing scheme and propose \model ~as well as the encoding, training, retraining and prediction methodology of ViT performance ranking. On the VIMER-UFO benchmark with 8 different computer vision tasks, \model ~is able to achieve comparable results yet achieve tremendous acceleration compared with baseline NAS algorithms. This paper departs from the conventional machine learning algorithm and provides a potentially new perspective to efficiently address the NAS problems, particularly those with limited architecture information available.


\begingroup
\setlength\bibitemsep{0pt}
\printbibliography
\endgroup

\end{document}